\documentclass[10pt,twocolumn,letterpaper]{article}

\usepackage{iccv}
\usepackage{times}
\usepackage{epsfig}
\usepackage{graphicx,float,diagbox,multirow}
\usepackage{amsmath}
\usepackage{amssymb}

\usepackage[breaklinks=true,bookmarks=false]{hyperref}

\iccvfinalcopy % *** Uncomment this line for the final submission

\def\iccvPaperID{****} % *** Enter the iccv Paper ID here

\ificcvfinal\fi

\begin{document}

%%%%%%%%% TITLE
\title{Temporal-Spatial Feature Pyramid for Video Saliency Detection}

\author{Qinyao Chang\\
Beihang University, \\
100191 Beijing, China.\\
{\tt\small changqinyao@buaa.edu.cn}
\and
Shiping Zhu\\
Beihang University, \\
100191 Beijing, China.\\
{\tt\small shiping.zhu@buaa.edu.cn}
}

\maketitle
\ificcvfinal\thispagestyle{empty}\fi

%%%%%%%%% ABSTRACT
\begin{abstract}
  Multi-level features are important for saliency detection. Better combination and use of multi-level features with time information can greatly improve the accuracy of the video saliency model. In order to fully combine multi-level features and make it serve the video saliency model, we propose a 3D fully convolutional encoder-decoder architecture for video saliency detection, which combines scale, space and time information for video saliency modeling. The encoder extracts multi-scale temporal-spatial features from the input continuous video frames, and then constructs temporal-spatial feature pyramid through temporal-spatial convolution and top-down feature integration. The decoder performs hierarchical decoding of temporal-spatial features from different scales, and finally produces a saliency map from the integration of multiple video frames. Our model is simple yet effective, and can run in real time. We perform abundant experiments, and the results indicate that the well-designed structure can improve the precision of video saliency detection significantly. Experimental results on three purely visual video saliency benchmarks and six audio-video saliency benchmarks demonstrate that our method outperforms the existing state-of-the-art methods.

\end{abstract}

%%%%%%%%% BODY TEXT

\section{Introduction}

Video saliency detection aims to predict the point of fixation for the human eyes while watching videos freely. It is widely applied in a lot of areas such as video compression~\cite{1hadizadeh2013saliency,2zhu2019high}, video surveillance~\cite{3guraya2010predictive,4yubing2011spatiotemporal} and video captioning~\cite{5nguyen2013static}.

Most of the existing video saliency detection models employ the encoder-decoder structure, and rely on the temporal recurrence to predict video saliency. For example, ACLNet~\cite{7wang2018revisiting} encodes static saliency features through attention mechanism, and then learns dynamic saliency through ConvLSTM~\cite{8shi2015convolutional}. SalEMA~\cite{9linardos2019simple} uses exponential moving average instead of LSTM to extract temporal features for video saliency detection. SalSAC~\cite{10wu2020salsac} proposes a correlation-based ConvLSTM to balance the alteration of saliency caused by the change of image characteristics of past frame and current frame. However, such a saliency modeling approach has the following problems. Firstly, the spatial saliency model is pretrained on the static image saliency datasets before finetuning on the video saliency datasets. However, the effectiveness of this transfer learning mechanism may be limited, since the resolutions of two datasets are different while saliency is greatly influenced by the image shape. Secondly, restricted by memory, the training of video saliency model requires to extract continuous video frames from the datasets randomly. However, the approach based on LSTM needs to utilize backpropagation through time to predict the video saliency of each frame. In this way, the state of LSTM of the first frame for the selected clip must be void, while, during the test, only the state of the LSTM of the first frame of the video is void, such discrepancy makes the modeling of method based on LSTM insufficient. Thirdly, as mentioned by~\cite{11min2019tased}, all the methods based on LSTM overlay the temporal information on top of spatial information, and fail to utilize both kinds of information at the same time, which is crucial for video saliency detection.
%\begin{figure}[h]
%\begin{center}
%   \includegraphics[width=\linewidth]{images/Fig1}
%\end{center}
%   \caption{The visualization of video saliency results of two different videos (interval of 30 frames)}
%\label{fig:1}
%\end{figure}

To alleviate above problems, some methods~\cite{11min2019tased, 37tsiami2020stavis, 28bellitto2020video} employ 3D convolutions to continuously aggregate the temporal and spatial clues of videos. While they achieve outstanding performance, there still remains an important issue, that is, lacking the utilization of multi-level features. Multi-level features are essential for the task of saliency detection, since the human visual mechanism is complicated and the concerned region is determined by various factors and from multiple levels. For example, some large objects may be salient, which are captured from the deeper layers with a relatively large receptive fields. Some small but moving at a high speed objects are also salient, which are captured from shallower layers holding more low-level information. Although the use of multi-level features such as FPN \cite{4422}  has already shined in the field of 2D object detection, there is currently few methods to fully verify that multi-level  features are effective for video saliency. [19] prove that multi-level features are effective for video saliency and achieve excellent performance. However, there is still room for research on how to better use and combine multi-level features and build a fully convolutional model to maximize the accuracy of the model.

In order to solve the past problems, we propose a new 3D fully convolutional encoder-decoder architecture for video saliency detection.
%, the generated saliency maps of video frames are shown in Figure~\ref{fig:1}.
 We  consider the influence of time, space and scale and establish a temporal-spatial feature pyramid similar to FPN \cite{4422}.
 %\cite{10wu2020salsac, 23lai2019video}. 
In this way, the temporal-spatial semantic features of deep layer are aggregated to each layer of the pyramid. In view of the different receptive fields of temporal dimension for the features of various layers, we separately perform independent hierarchical decoding on different levels of the feature pyramid to  take the effect of temporal-spatial saliency features with various scales into consideration. Some studies on the semantic segmentation of 2D network show that the convolution with the upsampling in decoder~\cite{15chen2014semantic,16chen2017deeplab,17chen2017rethinking,19chen2018encoder,18lin2017refinenet} can obtain better results, compared with some methods, which adopted the deconvolution or unpooling~\cite{14badrinarayanan2017segnet,12long2015fully,13ronneberger2015u}. We put away the previous deconvolution and unpooling operation of 3D fully convolutional encoder-decoder~\cite{11min2019tased} and completely adopt the 3D convolution and the trilinear upsampling.

At the same time, in order to predicting audio-video saliency, audio information and visual information are fused, and the obtained features are integrated in the original visual network in the form of attention. Our network is simple in structure, lower in parameters, and higher in prediction precision, which has obvious difference with other methods of the state-of-the-art. Our method ranks first in the largest and most diverse video saliency dataset, such as DHF1K~\cite{7wang2018revisiting}.

The main contributions of the paper are as following:

We develop a new 3D fully convolutional temporal-spatial feature pyramid network called TSFP-Net, which completely consists of 3D convolution and trilinear upsampling and obtain very high accuracy in the case of a small model size.

We construct feature pyramid of different scales containing rich temporal-spatial semantic features, and build a hierarchical 3D convolutional decoder to decode. We prove that such approach can significantly improve the detection performance of the video saliency. By fusing audio information and visual information and integrating them into the original visual network in the form of attention, we can simultaneously perform audio-video saliency prediction through TSFP-Net (with audio).

We evaluate our model on three purely visual large-scale video saliency datasets and six audio-video saliency datasets, comparing with the state-of-the-art methods, our model can achieve large gains.

\section{Ralated Work}

The video saliency detection consists of multiple directions, which mainly can be divided into two categories, fixation prediction and salient object detection. Fixation prediction aims to model the probability that the human eyes pay attention to each pixel while watching video images. We focus on the fixation prediction in this paper.

\subsection{The Latest 2D Video Saliency Detection Networks}

In the past, most video saliency detection methods predicted the saliency map by adding temporal recurrence module to the static network. DeepVS \cite{20jiang2018deepvs} establishes a sub-network of objects through YOLO~\cite{21redmon2016you} and builds up a sub-network of motion through FlowNet~\cite{22dosovitskiy2015flownet}, then, conveys the obtained spatial-temporal features to the double-layer ConvLSTM for prediction. ACLNet~\cite{7wang2018revisiting} adopts a attention module and a ConvLSTM module to construct the network, among which, the attention module is trained on the large static saliency dataset SALICON~\cite{34huang2015} and the ConvLSTM module is trained on the video saliency dataset. The final model is obtained through the alternating training of static and dynamic saliency. SalEMA~\cite{9linardos2019simple} discusses the performance of exponential moving average (EMA) and ConvLSTM for video saliency modeling and discovers that the former can acquire close or even better effect than ConvLSTM. STRA-Net~\cite{23lai2019video} proposes a kind of two-stream model, the motion flow and appearance can couple through dense residual cross-connections at various layers. Meanwhile, multiple local attentions can be utilized to enhance the integration of the temporal-spatial features and then conduct the final prediction of saliency map through ConvGRU and global attention. SalSAC~\cite{10wu2020salsac} improves the robustness of network through shuffled attention module, and the correlation-based ConvLSTM is employed to balance the change of static image feature for previous frame and current frame. ESAN-VSP~\cite{24chen2021video} adopts a multi-scale deformable convolutional alignment network (MDAN) to align the feature of adjacent frames and then predicts the video motion information through Bi-ConvLSTM. UNISAL~\cite{25droste2020unified} is a unified image and video saliency detection model, which can extract the static feature through MobileNet v2~\cite{35sandler2018mobilenetv2} and determine whether to predict the temporal information through the ConvGRU connected by the residual of the controllable switch. In addition, it also adopts the domain adaption technology to realize the high-precision saliency detection of various video datasets and image datasets.
\begin{figure*}[htbp]
\begin{center}
   \includegraphics[width=\linewidth]{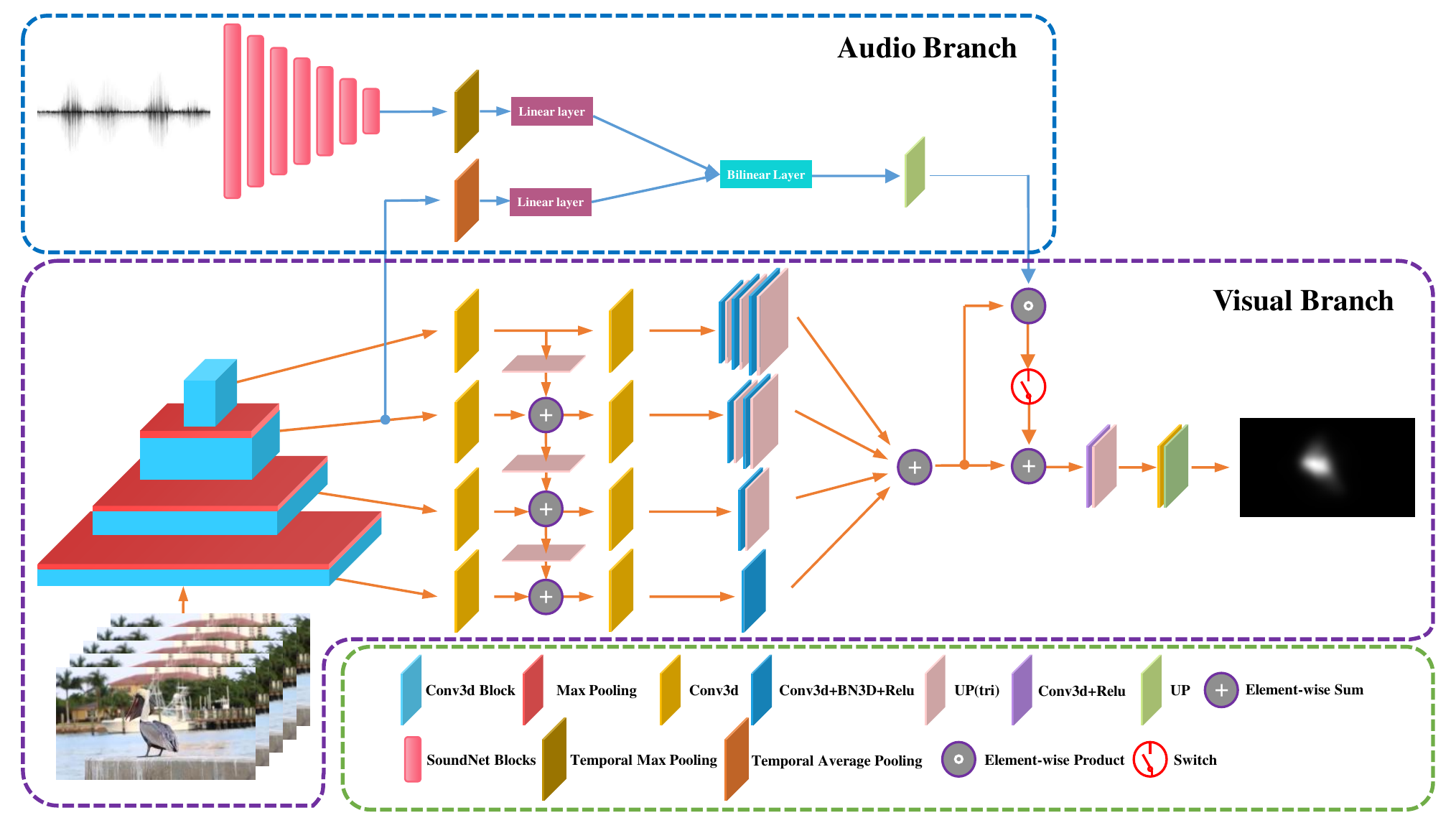}
\end{center}
   \caption{The overall architecture of the TSFP-Net (Notes: UP(tri) means trilinear upsampling, UP means bilinear upsampling)}
\label{fig:2}
\end{figure*}
\subsection{The Latest 3D Video Saliency Detection Networks}

RMDN~\cite{26bazzani2016recurrent} utilizes C3D~\cite{27tran2015learning} to extract the temporal-spatial features and then aggregates time information through LSTM. TASED-Net~\cite{11min2019tased} adopts S3D network~\cite{30xie2018} as encoder and the decoder uses 3D deconvolution and unpooling so as to  continuously enlarge the image to obtain the saliency map. The unpooling layer adopts {\itshape Auxiliary pooling} to fill the feature acquired from the decoder to the activated position corresponding to the maxpooling layer of the encoder. HD$^2$S~\cite{28bellitto2020video} delivers the multi-scale features output by 3D encoder to a conspicuity net for decoding separately and then combines all the decoded feature maps to obtain the final saliency map. ViNet~\cite{29jain2020avinet} adopts a 3D encoder-decoder structure in a 2D U-Net like fashion so that the decoding features of various layers can be constantly concatenated with the corresponding feature of encoder in the temporal dimension. And then, the video saliency detection results can be obtained through continuous 3D convolution and trilinear upsampling.

\subsection{Audio-Video Saliency Prediction}

Some recent studies have begun to explore the impact of the combination of vision and hearing on saliency. SoundNet~\cite{36aytar2016soundnet} uses a large amount of unlabeled sound data and video data, and uses a pre-trained visual model for self-supervised learning to obtain an acoustic representation. STAVIS~\cite{37tsiami2020stavis} performs a spatial sound source localization through SoundNet combined visual features in SUSiNet~\cite{38koutras2019susinet}, and concatenates the feature maps obtained through sound source localization and visual output feature maps to merge and output the saliency map. AViNet~\cite{29jain2020avinet} uses three different methods to fuse the advanced features of the SoundNet output with the deepest features of the ViNet encoder, and then performs audio-video saliency prediction.

We design a 3D fully convolutional encoder-decoder architecture for video saliency detection since the huge defect existed in the model designed in the 2D network described in the preceding part of the paper. Different from the above mentioned 3D network, our network completely utilizes the 3D convolutional layer and trilinear upsampling layer. Our network is the first to build temporal-spatial feature pyramid in the field of video saliency and aggregate deep semantic features in each layer of feature maps in the feature pyramid. Through the hierarchical decoding of temporal-spatial features at different scales, we obtain the detection results of video saliency that are superior to existing networks. At the same time, in order to perform audio-video saliency detection, we fuse sound information and visual information for sound source localization, and connect residuals with the features of the original visual network in the form of attention and then obtain audio-video saliency model.
\vspace{-0.5cm}
\section{The Proposed Methods}

\subsection{Network Structure}

The overall architecture of our model is shown in Figure~\ref{fig:2}. For purely visual branch TSFP-Net, since the saliency of any frame is determined by several frames in the past, hence, the network inputs $T$ frames at one time, and finally outputs a saliency map of the last frame of a $T$ frames video clip. That is, given the input video clip $\{I_{t-T+1}, \ldots, I_t\}$, the S3D \cite{30xie2018} encoder performs temporal-spatial feature aggregation through 3D convolution and maxpooling,  then, the top-down path enhancement integrates deep temporal-spatial semantic features into shallow feature maps  to establish the temporal-spatial feature pyramid, next, the temporal-spatial features with multi-scale semantic information are decoded hierarchically. The shallow features have smaller receptive fields, which are utilized to detect the small salient objects, and the deep layer features have the larger receptive fields, which are utilized to detect the large salient objects. As a result, the features of different levels are continuously decoded and upsampled so as to obtain the features with same temporal-spatial and channel dimension. These features are summed element by element, and the time and channel dimensions are reduced through the 3D convolution of the output layer. Finally, the saliency map $S_t$ at time $t$ is obtained through the sigmoid activation function.

For the combination of audio-video saliency, we input the sound waves of the T-frame clips corresponding to the visual network, obtaining the sound representation through SoundNet. We use the bilinear transformation to fuse the features of the base3 output of the S3D encoder with the sound representation, and then upsample to obtain the attention map. The attention map with the sound  information and the integrated visual feature map are combined by elementwise product and residual connection summation to obtain the features of the fused sound information. In this way, the visual features can be enhanced through sound information, and the residual connection ensures that the performance of the visual network can't be reduced. And the final audio-video saliency map is obtained through the output network.

In this way, in the form of a sliding window, each time we insert a new frame and delete the first frame, leaving the length of the video clip in the window as $T$. We can perform frame-by-frame video saliency detection, by doing so, all saliency results of the $T$ frames and subsequent frames of each video can be detected. For the first $T-1$ frames, we can obtain the saliency maps by roughly reversely playing the video frame of first $2T-1$ frames and putting them into the sliding window.

\subsection{Loss Function}

 In the past, a large number of video saliency models \cite{11min2019tased,29jain2020avinet} adopted {\itshape Kullback-Leibler (KL) divergence} as a loss function to train the model and achieved good results. However, there are multiple metrics that evaluate the saliency from different aspects, among them, the {\itshape Linear Correlation Coefficient (CC)} and the {\itshape Normalized Scanpath Saliency (NSS)} seem to be more reliable to evaluate the quality of the saliency map. We take the weighted summation of the above KL, CC and NSS to represent the final loss function and the subsequent ablation studies prove that the weighted summation of the three losses achieve better results than just using KL loss.

Assuming that the predicted saliency map is $S \in [0,1]$, the labeled binary fixation map is $F\in \{0,1\}$, and the ground truth saliency map generated by the fixation map is $G \in [0,1]$, the final loss function can be expressed as:
\[L(S,F,G) = L_{KL}(S,G) + \alpha_1 L_{CC}(S,G) + \alpha_2 L_{NSS}(S,F)\]
We set $\alpha_1 = 0.5$, $\alpha_2 = 0.1$ according to the value range of each item. $L_{KL}$, $L_{CC}$ and $L_{NSS}$ respectively signify the loss of {\itshape Kullback-Leibler (KL) divergence}, the {\itshape Linear Correlation Coefficient (CC)}, and the {\itshape Normalized Scanpath Saliency (NSS)}. The calculation formulas of them are as follows:
\begin{equation}
 L_{K L}(S, G)=\sum\nolimits_{x} G(x) \ln \frac{G(x)}{S(x)}
\end{equation}
\begin{equation}
 L_{C C}(S, G)=-\frac{cov(S, G)}{\rho(S) \rho(G)}
\end{equation}
\begin{equation}
\begin{aligned}
L_{N S S}(S, F)=&-\frac{1}{N} \sum\nolimits_{x} s(x) F(x),  \\
\Big(s(x)=&\frac{S(x)-\mu(S(x))}{\rho(S(x))}\Big)
\end{aligned}
\end{equation}
Where $\sum_x(\cdot)$ represents summing all the pixels, $cov(\cdot)$ represents the covariance, $\mu(\cdot)$ represents the mean and $\rho(\cdot)$ represents the variance.

\section{Experimental Results}

\subsection{Datasets}

Just like most video saliency studies, we evaluate our method on the three most commonly used video saliency datasets, which are DHF1K~\cite{7wang2018revisiting}, Hollywood-2~\cite{31mathe2014}, and UCF-sports~\cite{31mathe2014}. At the same time, we evaluate our model on six audio-video saliency datasets: DIEM~\cite{39mital2011clustering}, Coutrot1~\cite{40coutrot2014saliency}\cite{41coutrot2016multimodal}, Coutrot2~\cite{40coutrot2014saliency}\cite{41coutrot2016multimodal}, AVAD~\cite{42min2016fixation}, ETMD~\cite{43koutras2015perceptually}, SumMe~\cite{44gygli2014creating}.

\setcounter{table}{1}
\begin{table*}[htbp]
\setlength\tabcolsep{1.5em}
\begin{center}
\begin{tabular}{|c|c|c|c|c|c|}
\hline
\diagbox{Methods}{Metrics} & NSS & CC & SIM & AUC-J & s-AUC\\ \hline \hline
DeepVS~\cite{20jiang2018deepvs} & 1.911 & 0.344 & 0.256 & 0.856 & 0.583\\
ACLNet~\cite{7wang2018revisiting} & 2.354 & 0.434 & 0.315 & 0.890 & 0.601\\
SalEMA~\cite{9linardos2019simple} & 2.574 & 0.449 & \textcolor{red}{0.466} & 0.890 & 0.667\\
STRA-Net~\cite{23lai2019video} & 2.558 & 0.458 & 0.355 & 0.895 & 0.663\\
TASED-Net~\cite{11min2019tased} & 2.667 & 0.470 & 0.361 & 0.895 & 0.712\\
SalSAC~\cite{10wu2020salsac} & 2.673 & 0.479 & 0.357 & 0.896 & 0.697\\
UNISAL~\cite{25droste2020unified} & 2.776 & 0.490 & 0.390 & 0.901 & 0.691\\
HD$^2$S~\cite{28bellitto2020video} & 2.812 & 0.503 & \textcolor{blue}{0.406} & \textcolor{blue}{0.908} & 0.702\\
ViNet~\cite{29jain2020avinet} & \textcolor{blue}{2.872} & \textcolor{blue}{0.511} & 0.381 & \textcolor{blue}{0.908} & \textcolor{red}{0.729}\\
TSFP-Net & \textcolor{red}{2.966} & \textcolor{red}{0.517} & 0.392 & \textcolor{red}{0.912} & \textcolor{blue}{0.723}\\
\hline
\end{tabular}
\end{center}
\caption{Comparison of the saliency metrics on DHF1K test set for TSFP-Net and other state-of-the-art methods\ (The best scores are shown in red and second best scores in blue).}\label{tab2}
\end{table*}

\subsection{Experimental Setup}

In order to train TSFP-Net, we first initialize our encoder using the S3D model pre-trained on Kinetics. In the DHF1K dataset, we adopt standard division of training set and validation set to train our model. $T$ continuous video frames are randomly selected from each video in each time, each frame is resized to 192$\times$352, the batchsize is set to 16 videos during the training. Restricted by the memory, we can only deal with 4 videos each time, so we accumulate the gradient and update the model parameters every other 4 steps. We use the Adam optimizer, the initial learning rate is set to 0.0001, and the learning rate is reduced by 10 times at the 22nd, 25th, and 26th epochs respectively. We train 26 epochs in total, and use early stopping in the DHF1K validation set to save the model parameters corresponding to the largest NSS result on the validation set. Due to the excessive number of images in the validation set, we only use the first 80 frames of each video for validation during the training process.

As for Hollywood-2 and UCF-sports datasets, we use the models trained on DHF1K to finetune the models separately. Since these two datasets contain a large amount of video clips that are less than $T$, for all video clips less than $T$ in the training set, we first repeat the first frame $T-1$ times in front, and we adopt early stopping on the test set of these two datasets.

In order to train TSFP-Net (with audio), we first use the model pre-trained in DHF1K to initialize the visual branch and finetune on six audio-video saliency datasets without adding sound, and then add sound data to train the audio-video saliency model. The three different splits  used in the datasets are the same as~\cite{37tsiami2020stavis}, and we evaluate the average metrics of different splits.

{\bf Evaluation metrics.} We use the most commonly used metrics in the DHF1K benchmark to evaluate our model for DHF1K dataset. These include (i) Normalized Scanpath Saliency (NSS), (ii) Linear Correlation Coefficient (CC), (iii) Similarity (SIM), (iv) Area Under the Curve by Judd (AUC-J), and (v) Shuffled-AUC (s-AUC). For all these metrics, the larger, the better. For other datasets and ablation studies, we use AUC-J, SIM, CC and NSS metrics.

\subsection{Evaluation on DHF1K}

The DHF1K dataset is currently the largest and the most diverse video saliency dataset, thus, DHF1K is adopted as the preferred dataset for ablation study and evaluation of test set. We change the length of $T$ to 16, 32, and 48 respectively to train our model and observe the results on the DHF1K validation set. The experimental results are shown in Table~~\ref{tab1}. We discover that when $T$ is 32, the performance is the best, because it obtains the highest AUC-J, CC and NSS.

\setcounter{table}{0}
\begin{table}[htbp]
\begin{center}
\begin{tabular}{|c|c|c|c|c|}
\hline
Clip length ($T$) & AUC-J & SIM & CC & NSS\\ \hline \hline
16 & 0.916 & 0.392 & 0.500 & 2.876\\
32 & {\bf 0.919} & 0.397 & {\bf 0.529} & {\bf 3.009}\\
48 & 0.917 & {\bf 0.398} & 0.526 & 2.990\\
\hline
\end{tabular}
\end{center}
\caption{The experimental results of DHF1K validation set while training at different clip length ($T$).}\label{tab1}
\end{table}

Next, we submit the results of our model to the evaluation server of DHF1K test set. The results for TSFP-Net and all other state-of-the-art methods~\cite{28bellitto2020video,25droste2020unified,29jain2020avinet,20jiang2018deepvs,23lai2019video,9linardos2019simple,11min2019tased,7wang2018revisiting,10wu2020salsac} on DHF1K test set are shown in Table~~\ref{tab2}. We discover that our model is significantly better than other state-of-the-art methods, especially, NSS, CC and AUC-J make remarkable gains.  In particular, according to~\cite{33bylinskii2018},  NSS and CC are believed to be related to human eye’s visual attention most and recommended to evaluate the saliency model~\cite{33bylinskii2018}, compared with other methods, we make a huge breakthrough in terms of NSS and CC. Meanwhile, as known as in Table~~\ref{tab2}, the models based on 3D fully convolutional encoder-decoder are mostly superior to the 2D models based on LSTM~\cite{25droste2020unified,20jiang2018deepvs,23lai2019video,9linardos2019simple,7wang2018revisiting,10wu2020salsac}, which is related to the defects of the 2D network that we analyzed previously and the simultaneous temporal-spatial aggregation of 3D convolution. Our model is currently the most powerful 3D full convolutional encoder-decoder and video saliency network so far, which proves the effectiveness of our method.
 \setcounter{table}{3}
  \begin{table*}[htbp]
\begin{center}
%\small
\setlength\tabcolsep{.8em}
%\small
%\begin{spacing}{0.9}
\begin{tabular}{|c|c|c|c|c|c|c|c|c|}
\hline
\multirow{2}{*}{\diagbox{Methods}{Datasets}} & \multicolumn{4}{c|}{Hollywood-2} & \multicolumn{4}{c|}{UCF-sports}\\ \cline{2-9}
 & AUC-J & SIM & CC & NSS & AUC-J & SIM & CC & NSS\\ \hline\hline
DeepVS~\cite{20jiang2018deepvs} & 0.887 & 0.356 & 0.446 & 2.313 & 0.870 & 0.321 & 0.405 & 2.089\\
ACLNet~\cite{7wang2018revisiting} & 0.913 & 0.542 & 0.623 & 3.086 & 0.897 & 0.406 & 0.510 & 2.567\\
SalEMA~\cite{9linardos2019simple} & 0.919 & 0.487 & 0.613 & 3.186 & 0.906 & 0.431 & 0.544 & 2.638\\
STRA-Net~\cite{23lai2019video} & 0.923 & 0.536 & 0.662 & 3.478 & 0.910 & 0.479 & 0.593 & 3.018\\
TASED-Net~\cite{11min2019tased} & 0.918 & 0.507 & 0.646 & 3.302 & 0.899 & 0.469 & 0.582 & 2.920\\
SalSAC~\cite{10wu2020salsac} & 0.931 & 0.529 & 0.670 & 3.356 & \textcolor{red}{0.926} & \textcolor{blue}{0.534} & 0.671 & 3.523\\
UNISAL~\cite{25droste2020unified} & \textcolor{blue}{0.934} & 0.542 & 0.673 & \textcolor{blue}{3.901} & 0.918 & 0.523 & 0.644 & 3.381\\
HD$^2$S~\cite{28bellitto2020video} & \textcolor{red}{0.936} & \textcolor{blue}{0.551} & 0.670 & 3.352 & 0.904 & 0.507 & 0.604 & 3.114\\
ViNet~\cite{29jain2020avinet} & 0.930 & 0.550 & \textcolor{blue}{0.693} & 3.730 & \textcolor{blue}{0.924} & 0.522 & \textcolor{blue}{0.673} & \textcolor{blue}{3.620}\\
TSFP-Net & \textcolor{red}{0.936} & \textcolor{red}{0.571} & \textcolor{red}{0.711} & \textcolor{red}{3.910} & 0.923 & \textcolor{red}{0.561} & \textcolor{red}{0.685} & \textcolor{red}{3.698}\\ \hline
\end{tabular}
%\end{spacing}
\end{center}
\caption{The comparison of saliency metrics for TSFP-Net and other state-of-the-art methods on Hollywood-2 test set and  UCF-sports test set\ (The best scores are shown in red and second best scores in blue).}\label{tab4}
\end{table*}
  \begin{table*}[htbp]
\begin{center}
\setlength\tabcolsep{.2em}
%\resizebox{\textwidth}{12mm}
{
\begin{tabular}{|c|c|c|c|c|c|c|c|c|c|c|c|c|}
\hline
\multirow{2}{*}{\diagbox{Methods}{Datasets}} & \multicolumn{4}{c|}{DIEM} & \multicolumn{4}{c|}{Coutrot1} & \multicolumn{4}{c|}{Coutrot2}\\ \cline{2-13}
 & AUC-J & SIM & CC & NSS & AUC-J & SIM & CC & NSS & AUC-J & SIM & CC & NSS\\ \hline \hline
ACLNet~\cite{7wang2018revisiting} & 0.869 & 0.427 & 0.522 & 2.02 & 0.850 & 0.361 & 0.425 & 1.92 & 0.926 & 0.322 & 0.448 & 3.16\\
TASED-Net~\cite{11min2019tased} & 0.881 & 0.461 & 0.557 & 2.16 & 0.867 & 0.388 & 0.479 & 2.18 & 0.921 & 0.314 & 0.437 & 3.17\\
STAVIS~\cite{37tsiami2020stavis} & 0.883 & 0.482 & 0.579 & 2.26 & 0.868 & 0.393 & 0.472 & 2.11 & 0.958 & 0.511 & 0.734 & 5.28\\
ViNet~\cite{29jain2020avinet} & 0.898 & 0.483 & 0.626 & 2.47 & 0.886 & 0.423 & 0.551 & 2.68 & 0.950 & 0.466 & 0.724 & 5.61\\
AViNet(B)~\cite{29jain2020avinet} & 0.899 & 0.498 & 0.632 & 2.53 & 0.889 & 0.425 & 0.560 & 2.73 & 0.951 & 0.493 & \bf{0.754} & \bf{5.95}\\
TSFP-Net & 0.905 & \bf{0.529} & 0.649 & \bf{2.63} & 0.894 & \bf{0.451} & 0.570 & \bf{2.75} & 0.957 & 0.516 & 0.718 & 5.30\\
TSFP-Net (with audio) & \bf{0.906} & 0.527 & \bf{0.651} & 2.62 & \bf{0.895} & 0.447 & \bf{0.571} & 2.73 & \bf{0.959} & \bf{0.528} & 0.743 & 5.31\\ \hline
\end{tabular}}
\end{center}
\caption{Comparison results on the DIEM, Coutrot1 and Coutrot2 test sets (bold is the best).}\label{tab5}
\end{table*}%

We also compare the runtime  and the model size of our model with other state-of-the-art methods. We test our model on a NVIDIA RTX 2080 Ti, which take about 0.011s to generate a saliency map. The comparison of running time and model size with other methods is shown in Table~~\ref{tab3}. As can be seen that not only the accuracy of 
our model greatly exceed the state-of-the-art methods, but also the speed of generating the saliency map is fast enough, the model size is small enough but enough to obtain the highest accuracy.
\setcounter{table}{2}
\begin{table}[H]
\begin{center}
%
%\begin{spacing}{0.9}
\begin{tabular}{|c|c|c|}
\hline
Methods & Runtime (s)&Model Sizes (MB)\\ \hline \hline
DeepVS~\cite{20jiang2018deepvs} & 0.05&344\\
ACLNet~\cite{7wang2018revisiting} & 0.02&250\\
SalEMA~\cite{9linardos2019simple} & 0.01&364\\
STRA-Net~\cite{23lai2019video} & 0.02&641\\
TASED-Net~\cite{11min2019tased} & 0.06&82\\
SalSAC~\cite{10wu2020salsac} & 0.02&93.5\\
UNISAL~\cite{25droste2020unified} & 0.009&15.5\\
HD$^2$S~\cite{28bellitto2020video} & 0.03&116\\
ViNet~\cite{29jain2020avinet} & 0.016&124\\
TSFP-Net & 0.011&58.4\\
\hline
\end{tabular}
%\end{spacing}
\end{center}
\caption{The comparison of running time and model size for TSFP-Net and other state-of-the-art methods.}\label{tab3}
\end{table}

\vspace{-0.55cm}
\subsection{Evaluation on Other Datasets}

We evaluate the performance of our model on Hollywood-2 and UCF-sports. 
 The comparison results of our method on the Hollywood-2 and UCF-sports test sets and other state-of-the-art methods are shown in Table~\ref{tab4}, it can be seen that our model is also highly superior to other methods on these two datasets.

 We also evaluated the results of TSFP-Net and TSFP-Net (with audio) on six audio-video saliency datasets, and the performance comparisons with other methods are shown in Table~\ref{tab5} and Table~\ref{tab6}. It can be seen that our two models are much better than all the state-of-the-art methods on most datasets. We find that the model combined with sound has similar effect compared with pure visual model. We also try other fusion methods and find that the performance of the model doesn't improved or even poor. We believe that one possible reason is that the visual network is already sufficient to learn the saliency of very high precision on the existing datasets, so sound information is not needed. And the other possible reason is that the sound information is useless to the video saliency with the existing datasets, may need to explore larger and more diverse audio-video saliency datasets.

\setcounter{table}{5}
  \begin{table*}[htbp]
\begin{center}

\setlength\tabcolsep{.13em}
%\resizebox{\textwidth}{12mm}
{
\begin{tabular}{|c|c|c|c|c|c|c|c|c|c|c|c|c|}
\hline
\multirow{2}{*}{\diagbox{Methods}{Datasets}} & \multicolumn{4}{c|}{AVAD} & \multicolumn{4}{c|}{ETMD} & \multicolumn{4}{c|}{SumMe}\\ \cline{2-13}
 & AUC-J & SIM & CC & NSS & AUC-J & SIM & CC & NSS & AUC-J & SIM & CC & NSS\\ \hline \hline
ACLNet~\cite{7wang2018revisiting} & 0.905 & 0.446 & 0.58 & 3.17 & 0.915 & 0.329 & 0.477 & 2.36 & 0.868 & 0.296 & 0.379 & 1.79\\
TASED-Net~\cite{11min2019tased} & 0.914 & 0.439 & 0.601 & 3.16 & 0.916 & 0.366 & 0.509 & 2.63 & 0.884 & 0.333 & 0.428 & 2.1\\
STAVIS~\cite{37tsiami2020stavis} & 0.919 & 0.457 & 0.608 & 3.18 & 0.931 & 0.425 & 0.569 & 2.94 & 0.888 & 0.337 & 0.422 & 2.04\\
ViNet~\cite{29jain2020avinet} & 0.928 & 0.504 & 0.694 & \bf{3.82} & 0.928 & 0.409 & 0.569 & 3.06 & \bf{0.898} & 0.345 & \bf{0.466} & 2.40\\
AViNet(B)~\cite{29jain2020avinet} & 0.927 & 0.491 & 0.674 & 3.77 & 0.928 & 0.406 & 0.571 & 3.08 & 0.897 & 0.343 & 0.463 & \bf{2.41}\\
TSFP-Net & 0.931 & \bf{0.530} & 0.688 & 3.79 & \bf{0.932} & \bf{0.433} & \bf{0.576} & \bf{3.09} & 0.894 & \bf{0.362} & 0.463 & 2.28\\
TSFP-Net (with audio) & \bf{0.932} & 0.521 & \bf{0.704} & 3.77 & \bf{0.932} & 0.428 & \bf{0.576} & 3.07 & 0.894 & 0.360 & 0.464 & 2.30\\ \hline
\end{tabular}}
\end{center}
\caption{Comparison results on the AVAD, ETMD and SumMe test sets (bold is the best).}\label{tab6}
\end{table*}
  \begin{table*}[htbp]
\begin{center}
\setlength\tabcolsep{1.7em}
%\resizebox{\textwidth}{12mm}
{
\begin{tabular}{|c|c|c|c|c|c|c|c|c|c|c|c|c|}
\hline 
Methods& AUC-J & SIM & CC & NSS \\
\hline \hline
TSFP-Net (with audio) (zero audio) & $0.895$ & $0.446$ & $0.570$ & $2.73$ \\
TSFP-Net (with audio) & $0.895$ & $0.447$ & $0.571$ & $2.73$ \\
\hline

\end{tabular}}
\end{center}
\caption{Comparison of metrics of models with or without audio on the Coutrot1 dataset.}\label{newtab7}
%\end{table*}
%  \begin{table*}[htbp]
\begin{center}
\setlength\tabcolsep{1.8em}
%\footnotesize
\begin{tabular}{|c|c|c|c|c|}
\hline
Different Architecture & NSS & CC & AUC-J & SIM\\ \hline \hline
TSFP-Net (only final-level) & 2.7868 & 0.5010 & 0.9121 & 0.3860\\
TSFP-Net (only multi-level) & 2.8857 & 0.5097 & 0.9156 & 0.3819\\
TSFP-Net & {\bf 3.0086} & {\bf 0.5290} & {\bf 0.9188} & {\bf 0.3975}\\
\hline
\end{tabular}
\end{center}
\caption{Performance comparison for TSFP-Net with different network structures on the validation set of DHF1K.}\label{tab7}
%\end{table*}
%  \begin{table*}[htbp]
\begin{center}
\setlength\tabcolsep{1.9em}
\begin{tabular}{|c|c|c|c|c|}
\hline
Different Loss & NSS & CC & AUC-J & SIM\\ \hline \hline
TSFP-Net (only KL loss) & 2.9876 & 0.5287 & 0.9186 & 0.3927\\
TSFP-Net & {\bf 3.0086} & {\bf 0.5290} & {\bf 0.9188} & {\bf 0.3975}\\
\hline
\end{tabular}
\end{center}
\caption{Performance comparison for TSFP-Net with different loss functions on the validation set of DHF1K.}\label{tab8}
\end{table*}

We conduct a simple experiment to determine whether the sound is useful. We set the sound signal to a zero vector to observe the effect of the model when there is no sound, and compare it with the original model on the Coutrot1 dataset. The results obtained are shown in Table~\ref{newtab7}. It can be seen from the table that whether there is a sound signal or not does not affect the effect of the model. The difference in the experimental results comes from the jitter when the training data is sampled. Whether sound is useful for video saliency still needs to be explored, researching and establishing a larger-scale and more relevant audio-video saliency dataset for experiments is future research work.

\subsection{Ablation Studies}

We first prove that the multi-scale temporal-spatial feature pyramid constructed by top-down path enhancement and hierarchical decoding are effective and important for video saliency prediction.

Firstly, we only use the hierarchical decoder and do not build the temporal-spatial feature pyramid, we only change the channel dimensions of the output multi-scale temporal-spatial features through 1$\times$1$\times$1 convolution to make the feature channels input into the hierarchical decoder consistent. After that, the features directly input the hierarchical decoder for decoding and are integrated to obtain the saliency map, this configuration is TSFP-Net (only multi-level). Secondly, we delete the hierarchical decoder and only adopt the deepest features of the encoder for decoding to get saliency, the configuration is TSFP-Net (only final-level). The results on the validation set of DHF1K for different network structures are shown in Table~\ref{tab7}. We reveal that the results of hierarchical decoding for different layer's are significantly better than that obtained using only deepest layer’s features, and adding top-down path enhancement to construct a semantic temporal-spatial feature pyramid combined with hierarchical decoding has the best effect.

 Compared to TASED-Net~\cite{11min2019tased}, which adopts 3D deconvolution and unpooling, our TSFP-Net (only final-level) only adopts 3D convolution and trilinear upsampling. The NSS result on the validation set of DHF1K is 2.787, which is better than that of TASED-Net, which is 2.706. It indicates that deconvolution and unpooling not only rely too much on the maxpooling layer in the encoder, which leads to the inability to freely design the network structure, but also limits the learning ability of the network to some extent.
%

%\end{table}
%%
%%
%
%\begin{table}[htbp]

%

We also compare the effects of different loss functions on network performance, the results are shown in Table~\ref{tab8}. We prove that the adoption of the weighted summation of three losses can obtain better performance than using KL loss alone.

\section{Conclusion}

We propose a 3D fully convolutional encoder-decoder architecture to model the video saliency. Through the top-down path enhancement, we establish the multi-scale temporal-spatial feature pyramid with abundant semantic information. Then, the hierarchical 3D convolutional decoding is conducted to the multi-scale temporal-spatial features, and finally a video saliency detection model that is superior to all state-of-the-art methods is obtained. Our performances on three purely visual video saliency benchmarks and six audio-video saliency benchmarks prove the effectiveness of our method, and our model is real-time.

{\small
\bibliographystyle{ieee_fullname}
\bibliography{egbib}
}

\end{document}